\documentclass[letterpaper, 10 pt, conference]{ieeeconf}  % Comment this line out if you need a4paper

\IEEEoverridecommandlockouts                              % This command is only needed if 
\usepackage{
    amsbsy, amsfonts, amsmath, amssymb, bm, graphicx,
    multicol, mwe, siunitx, subcaption, xcolor, pdfpages, cuted, titlesec, float, booktabs, hyperref, balance
}
\let\labelindent\relax
\usepackage{enumitem}

\hypersetup{
    colorlinks=true,
    linkcolor=blue,
    }
\begin{document}

\title{\LARGE \bf
Prompt a Robot to Walk with Large Language Models
}

\author{Yen-Jen Wang$^{1,2,3}$, Bike Zhang$^{1}$, Jianyu Chen$^{2,3}$, Koushil Sreenath$^{1}$ % <-this % stops a space
\thanks{$^{1}$University of California, Berkeley. $^{2}$Institute for Interdisciplinary Information Sciences, Tsinghua University. $^{3}$Shanghai Qi Zhi Institute.}%
\thanks{This work is supported in part by the InnoHK of the Government of the Hong Kong Special Administrative Region via the Hong Kong Centre for Logistics Robotics and in part by The AI Institute. We thank Manfred Morari for discussions on a draft of this work.}
}

\maketitle
\thispagestyle{empty}
\pagestyle{empty}

%%%%%%%%%%%%%%%%%%%%%%%%%%%%%%%%%%%%%%%%%%%%%%%%%%%%%%%%%%%%%%%%%%%%%%%%%%%%%%%%
\begin{abstract}
Large language models (LLMs) pre-trained on vast internet-scale data have showcased remarkable capabilities across diverse domains. Recently, there has been escalating interest in deploying LLMs for robotics, aiming to harness the power of foundation models in real-world settings. However, this approach faces significant challenges, particularly in grounding these models in the physical world and in generating dynamic robot motions. To address these issues, we introduce a novel paradigm in which we use few-shot prompts collected from the physical environment, enabling the LLM to autoregressively predict low-level control actions for robots without task-specific fine-tuning. We utilize LLMs as a controller, diverging from the conventional approach of employing them primarily as planners. Simulation experiments across various robots and environments validate that our method can effectively prompt a robot to walk. We thus illustrate how LLMs can function as low-level feedback controllers for dynamic motion control, even in high-dimensional robotic systems. 
The project website and source code can be found at: \href{https://prompt2walk.github.io/}{prompt2walk.github.io}.
\end{abstract}

% However, current LLM drawbacks, including low-latency, limited context length, and high compute cost, preclude us from conducting hardware experiments.

%%%%%%%%%%%%%%%%%%%%%%%%%%%%%%%%%%%%%%%%%%%%%%%%%%%%%%%%%%%%%%%%%%%%%%%%%%%%%%%%
% \input{sections/introduction}
\section{INTRODUCTION}

\subsection{Motivation}
Large language models (LLMs) are foundational models that are pre-trained on internet-scale data \cite{brown2020language, radford2019language, radford2018improving, devlin2018bert, vaswani2017attention} and have demonstrated impressive results in various fields, such as natural language processing \cite{ouyang2022training, openai2023gpt}, computer vision \cite{radford2021learning}, and code generation \cite{chen2021evaluating}.
Recently, building upon the success of LLMs, there is a surging interest in utilizing LLMs for embodied agents \cite{ahn2022can, vemprala2023chatgpt}, aiming to harness the power of foundation models in the physical world \cite{bommasani2021opportunities}. Towards this goal, significant progress has been made in the form of robot foundation models \cite{brohan2022rt, brohan2023rt, driess2023palm}. However, such foundation models have to be specifically trained on large-scale robot-specific data that is not as easily available as textual data.
% previously performed low-level control, e.g., target joint positions, which were generally based on training or fine-tuning using significant volumes of robot data.

In this paper, we raise an intriguing question of whether off-the-shelf LLMs can function as low-level controllers for high-dimensional dynamical systems such as robots without any additional training.  While LLMs have been used to output high-level motion plans, the use of LLMs for low-level control is novel.  Our goal is to take a historical input-output sequence of a robotic system and get an LLM to output the next action to take and repeat this. 

\begin{figure}
    \centering
    \includegraphics[trim={0.0cm 3.2cm 0.0cm 0.0cm},clip,width=1.0\linewidth]{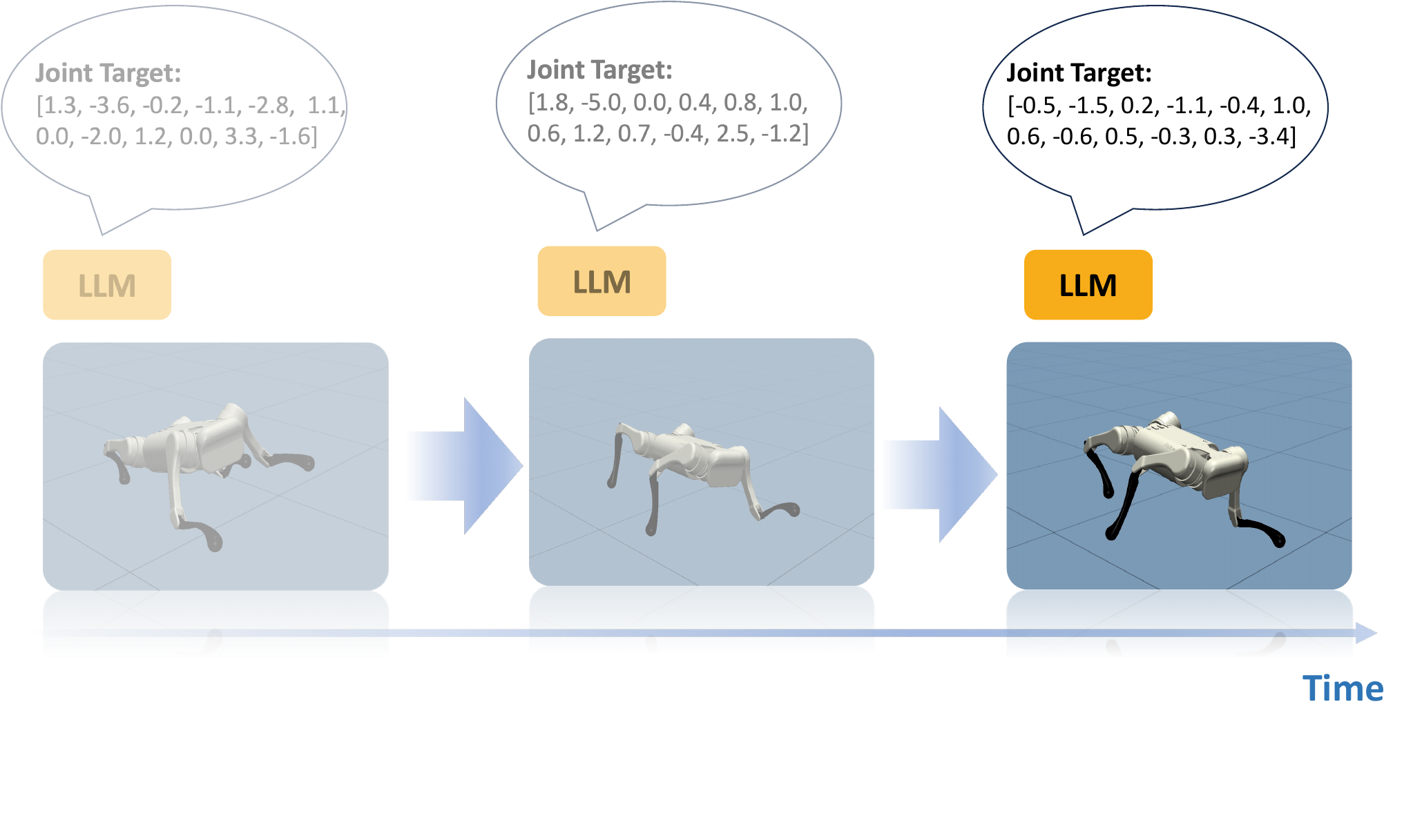}
    \caption{\textbf{Prompt a Robot to Walk.} Grounded in a physics-based simulator, LLMs output target joint positions to enable a robot to walk given a text prompt, which consists of a description prompt and an observation and action prompt.}
    \label{fig:cover}
    \vspace{-0.25in}
\end{figure}

\begin{figure*}
    \centering
    \includegraphics[width=0.8\linewidth]{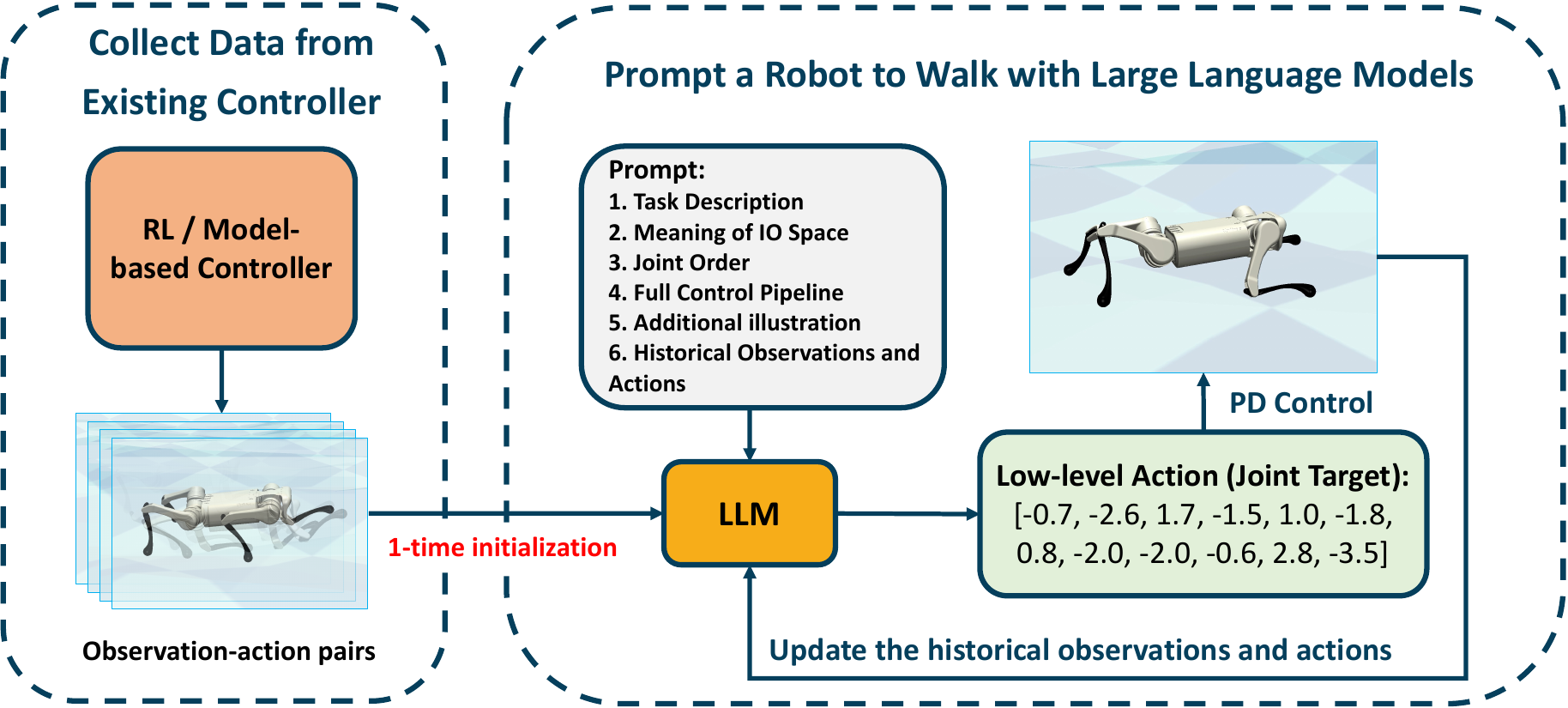}
    \caption{\textbf{LLM Policy Overview.} We first collect data from an existing controller to do a one-time initialization of the LLM prompt. Then, we design a text prompt including a description prompt and an observation and action prompt. The LLM outputs normalized target joint positions that are then tracked by a PD controller. After each LLM inference loop, the prompt is updated with the historical observations and actions. In our experiment, the LLM is supposed to run at $10$ Hz although the simulation has to be paused to wait for LLM inference, and the PD controller executes at $200$ Hz.}
    \label{fig:method}
    \vspace{-0.25in}
\end{figure*}

% \subsection{Introduction to LLMs for the Control Theoretician}
% \subsection{Background on Large Language Models}
% \subsection{Background on LLMs for the Control Theoretician}
\subsection{Background on LLMs}
We begin by defining various terms that are common in the field of LLMs but may not be familiar to someone in the controls community.

\textbf{Large Language Model (LLM).}
An LLM employs a transformer-based neural network, trained on extensive text data, to comprehend and produce human-like language, constructing information token by token—each representing a word part, single word, or phrase.
% An LLM utilizes a transformer-based deep neural network trained on vast amounts of text data, and is capable of understanding and generating human-like language. LLMs operate by generating information token by token, where each token could correspond to a part of a word, a single word or multiple words.

\textbf{Prompt.}
A prompt $P$ is a specific textual instruction or query given to the LLM to guide its language token generation~\cite{brown2020language}.  The process of modifying the prompt to improve the output is called prompt engineering.
The process of token generation of LLMs from a given prompt $P$ can be %succinctly 
described through the probabilistic model using a neural network~\cite{radford2018improving,radford2019language,ouyang2022training}: %as follows
% $P = \{p_{-k},...,p_{-1}\}$
%
% \begin{equation}
%     P_r(w_{n+1} | w_1, w_2, \ldots, w_n) = \frac{e^{s(w_{n+1}, w_1, w_2, \ldots, w_n)}}{\sum_{w'}e^{s(w', w_1, w_2, \ldots, w_n)}}
% \end{equation}
%
% Where $P_r(w_{n+1} | w_1, w_2, \ldots, w_n)$ is the conditional probability of the next word $w_n$ given the sequence of preceding words or tokens $w_{n+1}, w_1, w_2, \ldots, w_n \in P$ (the prompt), $s(w_{n+1}, w_1, w_2, \ldots, w_n)$ represents the score (e.g., logits before softmax) that the model assigns to the potential next word $w_{n + 1}$ given the preceding sequence.
%
\begin{equation}
    \emph{Pr}(w_{1}, w_{2}, \ldots, w_{T} | P) = \prod_{n=1}^{T} \frac{e^{s(w_{n} | w_1, w_2, \ldots, w_{n-1})}}{\sum_{w'}e^{s(w' | w_1, w_2, \ldots, w_{n-1})}},
\end{equation}
%
% \begin{equation}
%     \begin{aligned}
%     & h_0 = U W_e + W_p \\
%     & h_l = \text{transformer\_block}(h_{l-1}), \quad \forall i \in [1, n]  \\
%     & \emph{Pr}(u) = \text{softmax}(h_n W_e^T)
%     \end{aligned}
% \end{equation}
% where $U = \{u_{-j},...,u_{-1}\}$ is the context vector of tokens that are tokenized from prompt $P$, and $j$ is the context window. $W_e$ and $W_p$ are the token embedding matrix and the position embedding matrix, and $n$ is the number of transformer layers \cite{radford2018improving}.
%
where $T$ is the context length of the output, $w_{1}, w_{2}, \ldots, w_{T}$ are the output context, $\emph{Pr}(w_{1}, w_{2}, \ldots, w_{T} | P)$ is the conditional probability of the output given the prompt, $s(w_n | w_1, w_2, \ldots, w_{n - 1})$ represents a score %(e.g., logits before softmax) 
that the model assigns to the potential next word $w_n$ given the preceding sequence.

\textbf{Context of an LLM.}
The context of an LLM encompasses the input prompt and previous interactions, enabling it to generate relevant and informed responses. This dynamic context evolves with the conversation, guiding the LLM's understanding and output.

\textbf{Training LLMs.}
Training an LLM involves processing vast text data to learn language patterns, grammar, and nuances. This is achieved by adjusting the model's parameters during a pre-training phase, establishing its foundational language understanding.
% Training a LLM involves feeding a vast amount of text data into the model, allowing it to learn patterns, grammar, knowledge, and nuances of language from this data. The process involves adjusting the internal parameters of the model based on the input and expected output, a phase known as pre-training. This foundational phase sets the stage for the model's general understanding of language.

\textbf{Fine-tuning LLMs.}
Fine-tuning of an LLM involves adjusting its pre-trained weights using task-specific data, leading to improved performance on a particular task. This requires training the network further on new data tailored to the specific task.

% \textbf{In-Context Learning}
\textbf{Few-shot In-context Learning.}
In-context learning is a method of prompt engineering that enables LLMs to learn a new task from a few set of examples presented directly in the prompt. In particular, this happens without requiring any fine-tuning.

% \textbf{Zero-Shot vs Few-Shot In-Context Learning}
% Zero-shot learning enables LLMs to perform tasks they've never seen before, using only their pre-trained knowledge. Few-shot in-context learning improves this by providing a few examples to guide the LLM on new tasks. % We should link few-shot to in-context learning.
% : )

% \textbf{Aligning LLMs}
% Align LLMs ~= Ground LLM

\textbf{Grounding LLMs.}
%Grounding LLMs can enhance their understanding and generation capabilities by linking their outputs to real-world knowledge and context. %, facilitated by specific prompts provided.
%This integration is facilitated by tailored prompts.
Grounding refers to the process of linking the outputs of LLMs with real-world knowledge and context. This linkage enriches the understanding and generation capabilities of LLMs and is facilitated through tailored prompts.
%by providing them with tangible connections to the world around us. This integration of real-world context is facilitated by tailored prompts, which guide the LLMs to incorporate relevant information into their outputs, resulting in more accurate and contextually relevant responses.
% Grounding LLMs involves linking their outputs to real-world knowledge, enhancing their comprehension and generation capabilities. This integration is facilitated by tailored prompts.

\textbf{LLM as a Dynamical System.}
The evolution of a discrete-time dynamical system with input $u$, output $y$ and internal state $x$ can be written as
\begin{equation}
    x_{k+1} = f(x_k, u_k), \quad y_k = h(x_k, u_k),
\end{equation}
where the subscript $k$ refers to the $k^{th}$ time-step and $f, h$ represent the dynamics and output function of the dynamical system.  In a similar fashion, the $k^{th}$ interaction with an LLM with input prompt $P$, internal context $C$ and output $y$ can be captured as
\begin{equation}
    C_{k+1} = f_\theta(C_k, P_k), \quad y_k = h_\theta(C_k, P_k),
    \label{eq:LLM_dynamic_sys}
\end{equation}
where $f_\theta, h_\theta$ capture the context evolution and the output of the LLM at the $k^{th}$ interaction with the LLM.  Here the subscript $\theta$ captures the neural network parameters of the LLM that are obtained through training and remain fixed during inference / deployment.

\subsection{Contributions}
% We raise an intriguing question of whether LLMs can function as low-level controllers, which output the target joint position directly, for achieving dynamic tasks like robots walking through prompting. 

We explore a new paradigm that leverages few-shot prompts with an LLM, e.g., GPT-4, to output robot control actions, i.e., target joint position, directly. We utilize LLMs as a controller, diverging from the conventional approach of employing LLMs primarily as planners. We hypothesize that, given prompts collected from the physical environment, LLMs can learn to interact with it in-context, even though they are purely trained on text data. Moreover, we do not perform any fine-tuning of the LLM with task-specific robot data. We adopt a few-shot prompt approach, which contains historical observation and actions. %, and is widely adopted in the natural language field. 
Furthermore, we consider a dynamic control task of robot walking. 
A visualization of the paradigm is illustrated in Fig.~\ref{fig:cover}. We term this paradigm as \emph{prompting a robot to walk}. Grounded in a physical environment, the LLM takes  a designed text prompt, which includes a description prompt and an observation and action prompt, and outputs target joint positions to allow a robot to walk. Consequently, the robot is able to interact with the physical world through the generated control actions and gets the observations from the environment. In summary, the contributions are as follows:
\begin{itemize}
    \item Our main contribution is a framework for prompting a robot to walk with LLMs, where LLMs act as a feedback policy rather than a planner as is common in recent work.
    \item We propose and systematically analyze a text prompt design that enables LLMs to in-context learn robot walking behaviors.
    \item We extensively validate our framework on different robots, various terrains, and multiple simulators.
\end{itemize}
\section{Related Work}

\begin{figure*}
    \centering
    \includegraphics[trim={0.8cm 0.5cm 0.6cm 0.7cm},clip,width=0.85\linewidth]{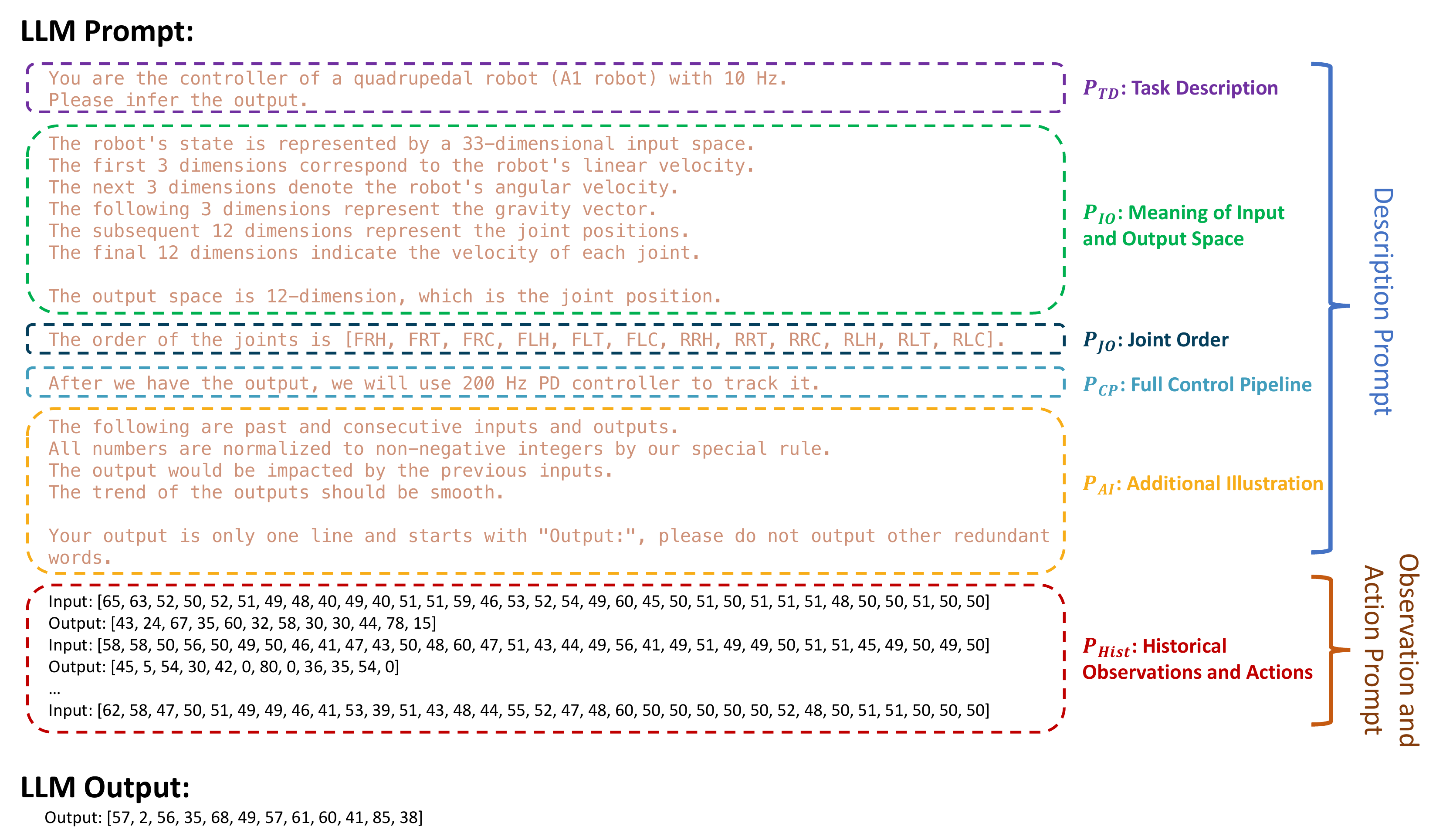}
    \caption{\textbf{Text Prompt.} We design a text prompt that includes two parts: a description prompt and an observation and action prompt. In the description prompt, we have the following subparts: $P_{TD}$: task description, $P_{IO}$: meaning of input and output space, $P_{JO}$: joint order, $P_{CP}$: full control pipeline, and $P_{AI}$: additional illustration. In the observation and action prompt, we have $P_{Hist}$: historical observations and actions. The LLM outputs normalized target joint positions.}
    \label{fig:prompt}
    \vspace{-0.2in}
\end{figure*}

\textbf{Large Language Models for Robotics.}
Large language models have recently become a popular tool for robotics including manipulation \cite{ahn2022can, driess2023palm, brohan2023rt, liang2023code, huang2022inner, yenamandra2023homerobot, huang2023voxposer}, locomotion \cite{tang2023saytap, yu2023language}, navigation \cite{huang2022language, shah2023lm, huang2023visual, guo2023doremi}, etc. Additionally, there are some recent research efforts to develop language agents \cite{yao2022react, sumers2023cognitive} using LLMs as the core.

With a focus on the intersection between LLMs and low-level robot control, \cite{wagener2022mocapact} trains a specialized GPT model using robot data to make a robot walk. However, our work directly uses the standard GPT-4 model without any fine-tuning. More interestingly, \cite{mirchandani2023large} instructs LLMs as general pattern machines and demonstrates a stabilizing controller for a cartpole in a sequence improvement manner \cite{zhou2022least}. Inspired by this work, we prompt LLMs to serve as a feedback policy for high-dimensional robot walking. Note that our work prompts a feedback policy without iterative improvement, whereas the cartpole controller in \cite{mirchandani2023large} is gradually improved as a return-conditioned policy. In addition, we explore textual descriptions to enhance the policy.

\textbf{Learning Robot Walking.}
Learning-based approaches have become promising methods to enable robots to walk. Deep reinforcement learning (RL) has been successfully applied to real-world robot walking \cite{tan2018sim, hwangbo2019learning}. In \cite{peng2020learning}, agile walking behavior is attained by imitating animals. To deploy a robot in complex environments, a teacher-student framework is proposed in \cite{lee2020learning, kumar2021rma}. Moreover, a robot can learn to walk in the real world \cite{smith2022legged, wu2023daydreamer}. Furthermore, the learning-based approach can enable dynamic walking behaviors \cite{xie2022glide, li2023learning, margolis2022rapid, yang2022fast, caluwaerts2023barkour, zhuang2023robot}.

More recently, LLMs have emerged as a useful tool for helping create learning-based policies for robot walking. In \cite{tang2023saytap}, contact patterns are instructed by human commands through LLMs. In \cite{yu2023language}, LLMs are utilized to define reward parameters for robot walking. In contrast to previous LLM-based robot walking work, we use LLMs to directly output low-level target joint positions.

\section{Method}
In this section, we present our method of prompting a robot to walk with large language models (LLMs). The overall framework is summarized in Fig.~\ref{fig:method}. We will first describe the data collection method to do a one-time initialization of the prompt, followed by our prompt engineering, and finally we will mention our approach on grounding the LLM.

% The LLMs are static feedback policies because the context changes between different calls, but nothing internal to the network, such as weights, changes.
% We start by introducing the data collection process from an existing controller. Then, we describe the prompt design for robot walking. Lastly, we illustrate how to ground LLMs for the robot walking task in a physics-based environment. 

% \begin{figure*}
%     \centering
%     \includegraphics[trim={0.8cm 0.5cm 0.6cm 0.7cm},clip,width=\linewidth]{figs/prompt.pdf}
%     \caption{\textbf{Text Prompt.} We design a text prompt that includes two parts: a description prompt and an observation and action prompt. In the description prompt, we have the following subparts: $P_{TD}$: task description, $P_{IO}$: meaning of input and output space, $P_{JO}$: joint order, $P_{CP}$: full control pipeline, and $P_{AI}$: additional illustration. In the observation and action prompt, we have $P_{Hist}$: historical observations and actions. The LLM outputs normalized target joint positions.}
%     \label{fig:prompt}
%     \vspace{-0.15in}
% \end{figure*}

\subsection{Data Collection}
A proper text prompt is one of the keys to utilizing LLMs for robot walking. We do one-time initialization of the prompt based on an existing controller, which could be either model-based or learning-based. From the existing controller, we collect observation and action pairs. The observation consists of sensor readings, e.g., IMU and joint encoders, while the action represents the target joint positions. It is important to note that the collected data serves as an initial input for LLM inference, whose output is then fed back to the LLM prompt. As the robot begins to interact with the environment and acquire new observations, the initial offline data will be replaced by LLM outputs. Thus, we consider this data collection phase as an initialization step.

\subsection{Prompt Engineering}
\label{sec:prompt_engineering}
Directly feeding observation and action pairs to LLMs often result in actions that do not achieve a stable walking gait. Next, we illustrate the prompt engineering step to guide LLMs in functioning as a feedback policy. Our prompt design, as shown in Fig.~\ref{fig:prompt}, can be classified into two categories: description prompt $P_{Desc}$ and historical observation and action prompt $P_{Hist}$ as below
\begin{equation}
    P = \{ \underbrace{P_{TD}, P_{IO}, P_{JO}, P_{CP}, P_{AI}}_{P_{Desc}}, P_{Hist} \}.
\end{equation}

% \begin{equation}
%     P = \{ P_{Desc}, P_{Hist} \}, \text{where} P_{Desc} = \{ P_{TD}, P_{IO}, P_{JO}, P_{CP}, P_{AI} \}.
% \end{equation}

\textbf{Description Prompt.}
\label{subsec:description_prompt}
The description prompt begins with $P_{TD}$, a precise task description of the robot walking task. This is then followed by control design details, e.g., the policy's operating frequency, ensuring that the LLM aligns the actions to this frequency. Next, we specify the format and meaning of both input observations and output actions in $P_{IO}$, allowing LLMs to understand the context of the inputs and actions. Then, an explicit enumeration of the joint order of our robot is provided in $P_{JO}$ to guide the LLM to comprehend the robot configuration. 
%Additionally, we specify in the prompt $P_{AI}$ to describe the data processing method and output requirements. Instead, these numerical values have been normalized. 
In $P_{AI}$, we provide additional illustration to describe the data processing method and output requirements.
Lastly, the prompt offers an overview of the entire control pipeline in $P_{CP}$, granting the LLM a macro perspective on how individual components enable it to process and interlink. It is crucial to highlight that, unlike classic learning-based and model-based walking controllers, text serves an important role in the LLM policy. 

\textbf{Observation and Action Prompt.}
\label{subsec:oa_prompt}
A sequence of observation and action pairs $P_{Hist}$ are used as prompts. These pairs are generated from the recent history of the robot walking trajectory. This procedure is widely used in RL-based robot walking controllers, where it allows the neural network to infer the dynamics as well as the privileged environment information. With a sequence of observation and action prompts, LLMs can in-context learn the dynamics and infer a reactive control action, where the observation prompt serves as the feedback signal. Note that both observation and action are converted to text format to interface with LLMs.

LLMs often struggle to comprehend the significance of numeric values, particularly floating point and negative numbers. Inspired by the prompt design in \cite{mirchandani2023large}, we adopt a normalization approach for numerical values. Specifically, we use a linear transformation to map all the potential numeric values into non-negative integers ranging from $0$ to $200$. We hypothesize that LLMs are mostly trained with text tokens. Thus, they are not sensitive enough to numerical values for robot control.

\subsection{Grounding LLMs}
In order to make LLMs useful for robot walking control, we need to ground them in a physical environment. We now introduce the pipeline to allow LLMs to interact with a robot and an environment. We use a physics-based simulator where LLMs can get observations from and send actions to. %The observations are from the physics-based simulation. 
The output of the LLM is the target joint positions, which are tracked by a set of joint Proportional-Derivative (PD) controllers running at a higher frequency. This joint-level PD control design is standard for learning-based robot walking control. While this pipeline is entirely done in simulation in this work, it has the potential to be implemented on hardware if the inference speed of LLMs is fast enough.

\section{Results}

\begin{figure}
    \centering
    \includegraphics[trim={0.4cm 0.4cm 0.4cm 0.4cm},clip,width=0.8\linewidth]{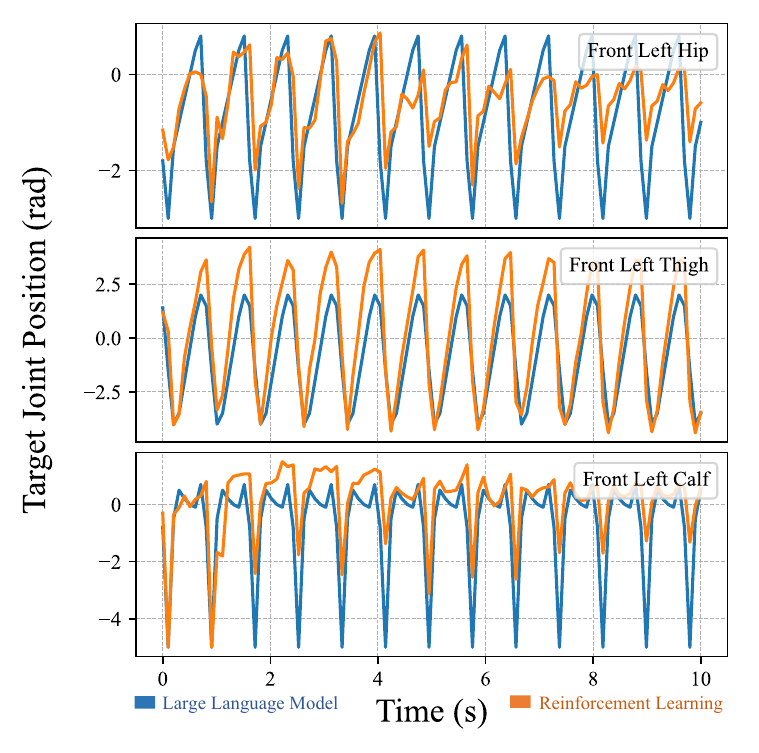}
    \caption{\textbf{Target Joint Position Trajectories.} The LLM and RL-based target joint position trajectories for the front left leg, including hip, thigh, and calf joints. The LLM trajectory is depicted in blue, and the RL trajectory is shown in orange.}
    \label{fig:joint_traj}
    \vspace{-0.15in}
\end{figure}

Having introduced the methodology for prompting a robot to walk, we next detail our experiments for validation. Moreover, through these experiments, we aim to answer the following questions:

\begin{enumerate}[label=Q\arabic*:]
    \item Can we prompt a robot to walk with LLMs?
    \item How should we design prompts for robot walking?
    \item Does the proposed approach generalize to different robots and environments?
\end{enumerate}

\subsection{Setup}
We choose an A1 quadruped robot as our testbed \cite{unitree}. It is a high-dimensional system with $12$ actuated joints. To initialize the LLM policy, we train an RL policy in Isaac Gym \cite{makoviychuk2021isaac} using Proximal Policy Optimization (PPO) \cite{schulman2017proximal}. This training is based on the training recipe from \cite{rudin2022learning}. Subsequently, we ground the LLM in Mujoco \cite{todorov2012mujoco}, a high-fidelity, physics-based simulator.
Our LLM policy operates at $10$ Hz \cite{gangapurwala2023learning} and is then tracked by a low-level joint PD controller at $200$ Hz. The P and D gains are set at $20$ and $0.5$, respectively. 

After evaluating various LLMs including GPT-4 \cite{openai2023gpt}, GPT-3.5-Turbo, text-davinci-003 \cite{openai_2023}, Alpaca \cite{taori2023alpaca}, Vicuna 2 \cite{chiang2023vicuna}, Llama 2 \cite{touvron2023llama}, we found that only GPT-4 is powerful enough to in-context learn a robot walking behavior using our designed prompt. During the experiments, we set GPT-4's temperature to $0$ to minimize the variance.

\begin{figure}
    \centering
    \includegraphics[trim={0.4cm 0.3cm 0.3cm 0.3cm},clip,width=0.75\linewidth]{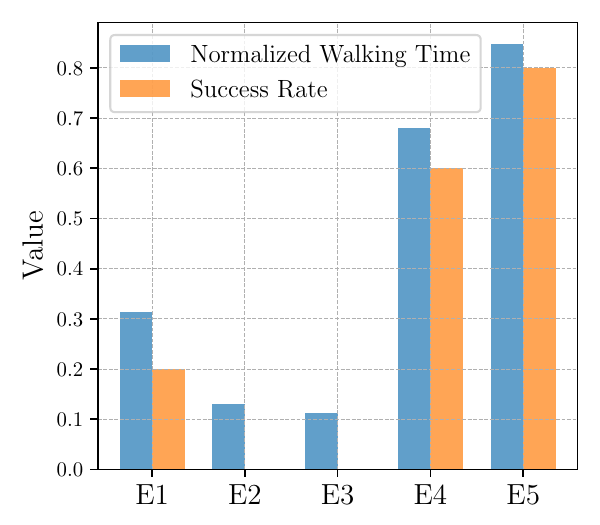}
    \caption{\textbf{Description Prompt Comparison.} (E1) No description prompt (i.e. only $P_{Hist}$.) (E2) $P_{Desc} = \{P_{IO}\}$: meaning of input and output space. (E3) $P_{Desc} = \{P_{IO}, P_{JO}\}$: meaning of input and output space and joint order. (E4) $P_{Desc} = \{P_{TD}, P_{IO}, P_{JO}, P_{CP}\}$: task description, meaning of input and output space, joint order, and full control pipeline. (E5) Full description prompt.}
    \label{fig:description_prompt}
    \vspace{-0.25in}
\end{figure}

\subsection{Robot Walking}
Utilizing the proposed approach, we successfully prompt an A1 quadruped robot to walk with GPT-4.
The LLM policy can not only enable walking on flat ground but can also allow the robot to walk over uneven terrain, as shown in Fig.~\ref{fig:anymal}. Due to the unexpected roughness, the robot almost falls over, but the LLM policy allows it to recover to a normal posture and then keep walking forward. Due to the need to balance the token limit of the LLM and the size of $P_{Hist}$, we execute the policy at $10$ Hz. However, this leads to a walking gait that becomes reasonably worse compared to many RL-based walking policies running at $50$ Hz or even higher.

Fig.~\ref{fig:joint_traj} demonstrates target joint trajectories for the front left leg when a robot is walking on uneven terrain for $10$ seconds. The blue lines depict the trajectories produced by the LLM policy. As a comparison, the orange lines show the trajectories generated by an RL policy. Note that both trajectories take the same observation as input. The robot acts with the action generated by the LLM and then gets the next observation from the environment. Although the LLM policy is initialized with the RL policy, the resulting joint trajectories are noticeably different.

One prompt example for A1 robot walking is shown in Fig.~\ref{fig:prompt}, where we use historical observations and actions for the past $50$ steps. The prompt is specially designed and normalized as described in Sec.~\ref{sec:prompt_engineering}. Based on this A1 robot walking experiment, we can answer Question Q1, which is that a robot can be prompted to walk with LLMs.

\subsection{Description Prompt}
We perform $5$ experiments to analyze the impact of individual components in the description prompt. In each experiment, we provide observation and action prompts ($P_{Hist}$). For evaluation, we consider two metrics: normalized walking time and success rate. To clarify, the term ``normalized walking time" denotes the proportion of time a robot can walk before it falls. The success rate is measured by the percentage of the trials that the robot is able to finish, where each trial lasts for $10$ seconds, and we have $5$ trials for each experiment. In the design of the first experiment (E1), we exclude the description prompt entirely (we only have $P_{Hist}$, and set $P_{Desc} = \emptyset$). In the second experiment (E2), we only provide the meaning of input and output space ($P_{Desc} = \{P_{IO}\}$). In the third experiment (E3), we include the joint order ($P_{Desc} = \{P_{IO}, P_{JO}\}$). In the fourth experiment (E4), we incorporate prompts such as task description, meaning of input and output space, joint order, and the full control pipeline ($P_{Desc} = \{P_{TD}, P_{IO}, P_{JO}, P_{CP}\}$). For the fifth experiment (E5), we employed a complete description prompt. The experimental result is demonstrated in Fig.~\ref{fig:description_prompt}, where we can see that the full description prompt has the highest normalized walking time and success rate. Based on the results from the first experiment, without a description prompt (E1), there is a minimal likelihood of LLMs prompting a robot to walk.

\begin{figure}
    \centering
    \includegraphics[trim={0.25cm 0.39cm 0.3cm 0.3cm},clip,width=0.8\linewidth]{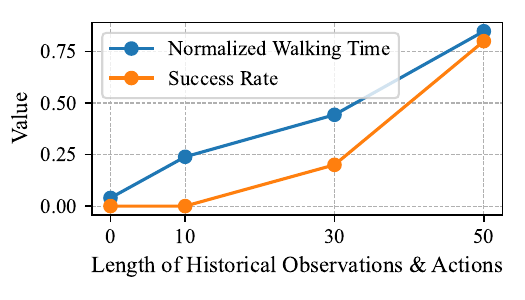}
    \caption{\textbf{Observation and Action Length Comparison.} We conduct experiments for historical observation and action lengths of size $0$, $10$, $30$, and $50$. With lengths ranging from $0$ to $50$, the LLM token consumption is approximately $348, 1738, 4518$, and $7298$ tokens, respectively.}
    \label{fig:obs_length}
    \vspace{-0.3in}
\end{figure}

\subsection{Observation and Action Prompt}
In our subsequent investigation, we assess the influence of the observation and action prompt $P_{Hist}$ on walking performance. Inspired by the RL-based walking control design, we first study how historical observations and actions affect the performance. We conduct a series of experiments, testing observation and action lengths of $0, 10, 30$, and $50$, all while using the description prompt. To clarify, a length of $0$ means only a description prompt. In our experiments, the LLM is queried at $10$ Hz, so a length of $50$ means that $P_{Hist}$ captures a 5-second history of observation and action that covers several walking steps for a quadruped robot. The experimental result is shown in Fig.~\ref{fig:obs_length}. It is evident that increased length of observations and actions correlate with enhanced performance, both in terms of normalized walking time and success rate. With lengths ranging from $0$ to $50$, the LLM token consumptions are approximately $348, 1738, 4518$, and $7298$, respectively. As we use the GPT-4 model with an 8k token length, we are not able to explore longer lengths of observations and actions.

\begin{figure}
    \centering
    \includegraphics[trim={0.4cm 0.3cm 0.3cm 0.3cm},clip,width=0.75\linewidth]{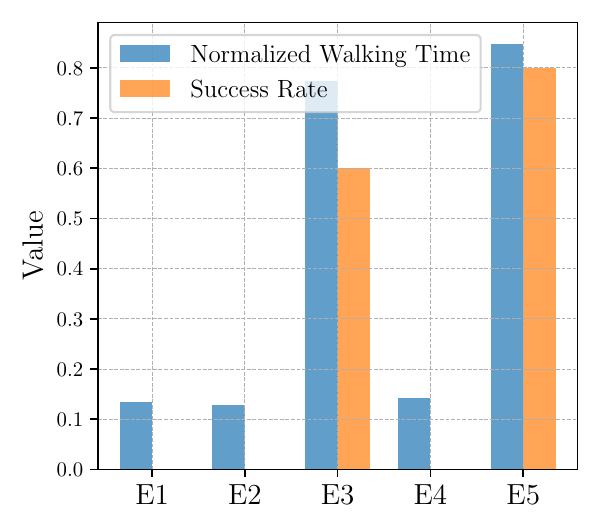}
    \caption{\textbf{Observation Choice Comparison.} (E1) No observation. (E2) Base linear velocity and angular velocity. (E3) Joint position and joint velocity. (E4) Combine observations from experiments 2 and 3. (E5) Full observation.}
    \label{fig:obs_choice}
    \vspace{-0.35in}
\end{figure}

In addition to comparing various lengths for observation and action prompts, we also investigate the effect of different observation prompts. Our choices for observations are influenced by the RL policy, as we initialize our LLM policy using a reinforcement learning-based approach. We evaluated five scenarios: (E1) no observation; (E2) only base linear velocity and angular velocity; (E3) only joint position and joint velocity; (E4) a combination of base linear velocity, angular velocity, joint position, and joint velocity; (E5) full observation. The comparison result is shown in Fig.~\ref{fig:obs_choice}. It is important to note that in the E1 experiment, only actions were provided, which essentially amounts to an open-loop control. This form of control was insufficient for successfully making the robot walk in the experiment. Intuitively, it seems right. The full observation listed in Fig.~\ref{fig:prompt} achieves the best performance. However, it remains unclear which specific observation component is the most influential. It is noteworthy that the LLM policy operates with an observation space of 33 dimensions, whereas the RL policy uses 48 dimensions. Since an observation space of 48 dimensions would require an excessive number of tokens for LLMs, we carefully selected 33 key dimensions essential for walking control, which proved sufficient for enabling the robot to walk. Although the LLM policy may not perform as well as RL in terms of precision, it demonstrates an ability to understand underlying physical principles from the trajectory generated by the RL policy. This allows the agent to move effectively in simulation, leveraging the insights gleaned from RL to guide its own control decisions.

Furthermore, we study the effect of how we normalize the observation and action prompt. We benchmark $5$ different normalization methods: (E1) original values without any normalization; (E2) normalize to positive values; (E3) normalize to integers; (E4) discard the decimal part and then normalize the integer part to positive integer values; (E5) normalize to positive integer values. Due to the limited token size of GPT-4, we opt for a compact observation prompt consisting of base linear and angular velocities. The benchmark result is summarized in TABLE~\ref{tab:normalization}. Unlike other experiments, to emphasize the performance in different normalization methods, we extend the walking time to $20$ seconds. We found that the normalization of the observation and action prompt is crucial as LLMs might parse a value of observation or action into several text tokens. In GPT-4, integers within the range of [-300, 300] are tokenized as a single token. The experimental results indicate that representing a single value as a token facilitates the LLM's ability to uncover implicit relationships between numbers.

Based on the investigation of the text prompt, we can answer Question Q2: how should we design prompts for robot walking? We believe a synergy between description prompt and observation and action prompt is the key to utilizing LLMs to prompt a robot to walk.

\subsection{Different Robots}
\begin{figure}
    \centering
    \includegraphics[trim={0.0cm 4.2cm 0.0cm 0.0cm},clip,width=0.8\linewidth]{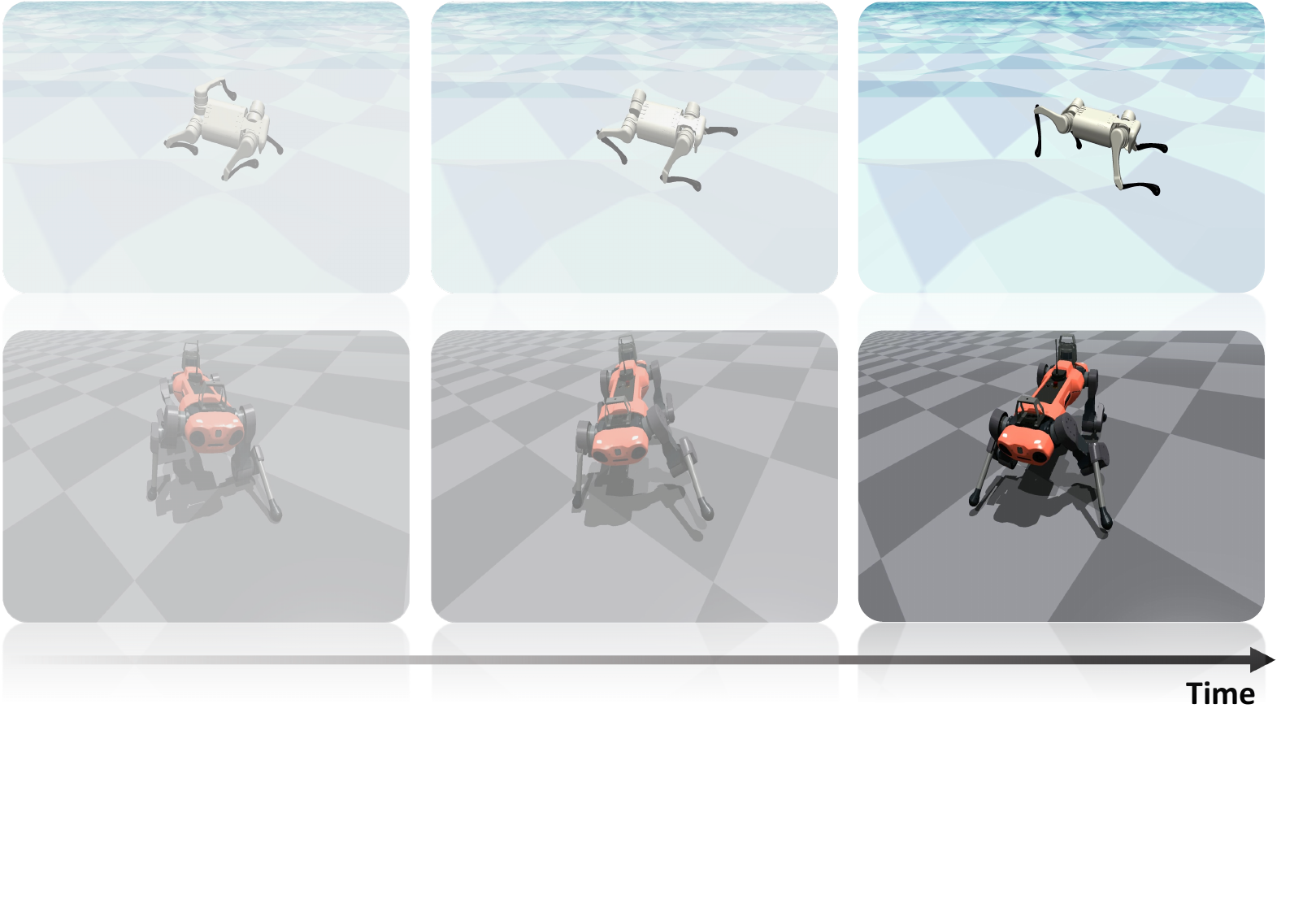}
    \caption{\textbf{Robot Walking Visualization.} Top: A1 robot is prompted to walk on uneven terrain in MuJoCo, where the LLM policy can make it recover from terrain disturbance. Bottom: ANYmal robot is prompted to walk on flat ground in Isaac Gym using the same approach.}
    \label{fig:anymal}
    \vspace{-0.25in}
\end{figure}

In addition to the A1 robot, we further validate our approach with a different robot: the ANYmal robot \cite{hutter2016anymal}. It is different from the A1 robot in terms of size, mass, mechanical design, etc. In this experiment, we use Isaac Gym instead of MuJoCo as our simulator to see the effect of change in the simulation environment. Following the same approach, we train a $10$ Hz RL policy for initialization. With the proposed text prompt, we successfully prompt the ANYmal robot to walk on flat ground. Snapshots of ANYmal walking are shown in Fig.~\ref{fig:anymal}. Having been validated by the A1 and ANYmal experiments over various terrains, we believe that the proposed method generalizes to different robots and environments, which is our answer to Question Q3.

\begin{table}
\begin{center}
\begin{tabular*}{0.91\linewidth}{@{}llcccc@{}}
\toprule
  Experiment & E1 & E2 & E3 & E4 & E5 \\
\midrule
  NWT(↑) [\%] & 0.137 & 0.086 & 0.700 & 0.504 & \textbf{0.721} \\
  Success Rate(↑) [\%] & 0.0 & 0.0 & \textbf{0.6} & 0.2 & \textbf{0.6} \\
  No. Input Tokens(↓) & 4947 & 5117 & \textbf{3135} & \textbf{3135} & \textbf{3135} \\
  No. Output Tokens(↓) & 62 & 62 & \textbf{38} & \textbf{38} & \textbf{38} \\
\bottomrule
\end{tabular*}
\caption{\textbf{Normalization Method Benchmark.} (E1) Original values. (E2) Normalize to positive values. (E3) Normalize to integer values. (E4) Discard the decimal and then normalize the integer to positive integer values. (E5) Normalize to positive integer values. NWT is normalized walking time.}
\label{tab:normalization}
\end{center}
\vspace{-0.35in}
\end{table}

\section{Discussion}

After validating our approach with experimental results, we discuss what we learned in this study and the limitations of the current approach.

\subsection{Text is Another Interface for Control}
It is interesting to note that the description prompt plays a crucial role in utilizing LLMs to prompt a robot to walk, which indicates that text is another interface for control. The existing control approaches for robot walking do not rely on any task description in textual form. If we follow the convention of RL or model-based control that uses numerical values such as observations and actions, LLMs have a low chance of making a robot walk, as demonstrated in Fig.~\ref{fig:description_prompt}. Instead, with a proper design of the description prompt, LLMs can achieve a high success rate for walking. We hypothesize that a description prompt provides a context for LLMs to interpret the observations and actions properly. While we provide a prompt example for robot walking, the prompt design for robot motions is still under-explored.

\subsection{LLMs In-Context Learn Differently}
Our experiments demonstrate that LLMs in-context learn to prompt a robot to walk. Initially, we hypothesized that LLMs might learn a robot walking behavior in a manner akin to behavior cloning~\cite{wei2023larger}. However, as shown in Fig.~\ref{fig:joint_traj}, the joint trajectories generated by the LLM policy are sufficiently different from those generated by an RL policy. Moreover, the LLM policy shows a more regular pattern, which is not present in the RL policy. If we pay attention to the left calf joint trajectory, the pattern coincides with the biomechanics study of animal walking \cite{holmes2006dynamics}. Thus, we believe that LLMs in-context learn differently to enable a robot to walk.

\subsection{LLMs as Dynamic Feedback Controllers}
In a typical neural network trained via reinforcement learning, the policy's action (neural-network output) is a function of its current state (neural-network input), acting as a static feedback controller. However, for LLMs, the output is influenced not just by the input prompt but also by the contextual state, which evolves with each new prompt and response, see \eqref{eq:LLM_dynamic_sys}. In this sense, we LLMs are dynamic feedback controllers. 
% Futhermore, due to the evolving contextual state, we also ... ?
% Furthermore, as the internal activation of LLMs can be changed based on the context,
Furthermore, due to the evolving contextual state, we also hypothesize that the LLM is similar to an adaptive controller that typically utilizes a history of previous system inputs and outputs to adjust its parameters.  % Why talk about internal activation - the context itself changes with the inputs and outputs.  The context is then like a state..  (the internal activation probably changes even for a NN with different inputs.)

% In a typical neural network (NN) policy that is trained through reinforcement learning, the action (NN output) of the learned policy is a function of the state (NN input) and is thus a static feedback controller.  While LLMs are also neural networks, the action (LLM output) is not only a function of the prompt (LLM input) but also a function of the state (LLM context).  Moreover, the context updates as new prompts are received and responses generated. In this sense, we hypothesize that LLMs are dynamic feedback controllers.

\subsection{Limitations}
While this work takes us closer to utilizing LLMs for robot walking control, there are some limitations in the current framework. First, the current prompt design is fragile. Minor alterations in the prompt can dramatically affect the walking performance, as described in our experiments. In general, we still lack a good understanding of how to design a reliable prompt for robot walking.
Secondly, as we design and test the prompt based on a specific initialization policy, our prompt design inevitably becomes biased toward this policy. Although we have tested our framework with several different RL initialization policies, it is possible that some initialization policies do not work with our prompt.

Another major limitation is that we are only able to carry out simulation experiments instead of hardware experiments. One reason is the low inference speed of GPT-4. Our pipeline requires LLMs to be queried at $10$ Hz, which is much faster than the actual inference speed through OpenAI API. Thus, we have to pause the simulation to wait for the output of GPT-4. Furthermore, due to the limited token size, we have to choose a low-frequency policy, i.e., $10$ Hz, to maximize the time horizon of the context. As a side note for future research, this work was expensive and roughly costed $\$2,000$ US dollars for all the OpenAI API calls.
\section{Conclusions}
In this paper, we presented an approach for prompting a robot to walk. We use LLMs with text prompts, consisting of a description prompt and an observation and action prompt collected from the physical environment, without any task-specific fine-tuning. As an early exploration of LLMs in the context of physics, our study investigates the feasibility of LLMs in understanding and interacting with the physical world in simulation. Our experiments demonstrate that LLMs can serve as low-level feedback controllers for dynamic motion control even in high-dimensional robotic systems. We further systematically analyzed the text prompt with extensive experiments. Furthermore, we validated this method across various robotic platforms, terrains, and simulators. In the future, we aim to address the current limitations of this work and refine our proposed method for application in real-world physical environments.

% \addtolength{\textheight}{-12cm}   % This command serves to balance the column lengths
                                  % on the last page of the document manually. It shortens
                                  % the textheight of the last page by a suitable amount.
                                  % This command does not take effect until the next page
                                  % so it should come on the page before the last. Make
                                  % sure that you do not shorten the textheight too much.

%%%%%%%%%%%%%%%%%%%%%%%%%%%%%%%%%%%%%%%%%%%%%%%%%%%%%%%%%%%%%%%%%%%%%%%%%%%%%%%%

%%%%%%%%%%%%%%%%%%%%%%%%%%%%%%%%%%%%%%%%%%%%%%%%%%%%%%%%%%%%%%%%%%%%%%%%%%%%%%%%

%%%%%%%%%%%%%%%%%%%%%%%%%%%%%%%%%%%%%%%%%%%%%%%%%%%%%%%%%%%%%%%%%%%%%%%%%%%%%%%%

% \newpage
\bibliographystyle{IEEEtranS}
\bibliography{references}{}

\balance

\end{document}